\documentclass[10pt,conference]{IEEEtran}
\IEEEoverridecommandlockouts

\usepackage{glossaries} 
\usepackage{hyperref}
\usepackage{graphicx}
\usepackage{cite}
\usepackage{amsmath,amssymb,amsfonts}
\usepackage[capitalise]{cleveref} 
\crefformat{equation}{(#2#1#3)} 
\usepackage{makecell}
\usepackage[caption=false,font=footnotesize]{subfig}

\newacronym{ransac}{RANSAC}{RANdom SAmple Consensus}
\newacronym{slam}{SLAM}{Simultaneous Localization and Mapping}
\newacronym{lidar}{LiDAR}{Light Detection And Ranging}
\newacronym{sota}{SOTA}{State-of-the-art}
\newacronym{icp}{ICP}{Iterative Closest Point}
\newacronym{mos}{MOS}{Moving Object Segmentation}
\newacronym{rtk}{RTK}{Realt-Time Kinematic}
\newacronym{gnss}{GNSS}{Global Navigation Satellite System}
\newacronym{datmo}{DATMO}{Detection And Tracking of Moving Objects}
\newacronym{cnn}{CNN}{Convolutional Neural Network}
\newacronym{rnn}{RNN}{Recurrent Neural Network}
\newacronym{ipm}{IPM}{Inverse Perspective Mapping}
\newacronym{ins}{INS}{Inertial Navigation System}
\newacronym{imu}{IMU}{Inertial Measurement Unit}
\newacronym{enc}{ENC}{Electronic Navigational Chart}
\newacronym{dbscan}{DBSCAN}{Density-Based Spatial Clustering of Applications with Noise}
\newacronym{radar}{RADAR}{Radio Detection And Ranging}
\newacronym{eot}{EOT}{Extended Object Tracking}
\newacronym{vjipda}{VJIPDA}{Visual Joint Integrated Probabilistic Data Association}

\begin{document}

\title{Near-Shore Mapping for Detection and Tracking of Vessels}

\author{
Nicholas Dalhaug, Annette Stahl, Rudolf Mester and Edmund Førland Brekke
\thanks{\raggedright This work was supported by The Research Council of Norway (project number 333917).}
\thanks{N. Dalhaug is with the Norwegian University of Science and Technology (NTNU), Trondheim, Norway, e-mail: nicholas.dalhaug@ntnu.no.}
\thanks{A. Stahl, R. Mester and E. F. Brekke are also with NTNU.}
}

\maketitle

\begin{abstract}
For an autonomous surface vessel (ASV) to dock, it must track other vessels close to the docking area. Kayaks present a particular challenge due to their proximity to the dock and relatively small size. Maritime target tracking has typically employed land masking to filter out land and the dock. However, imprecise land masking makes it difficult to track close-to-dock objects. 
Our approach uses \gls{lidar} data and maps the docking area offline. The precise 3D measurements allow for precise map creation. However, the mapping could result in static, yet potentially moving, objects being mapped. We detect and filter out potentially moving objects from the \gls{lidar} data by utilizing image data. 
The visual vessel detection and segmentation method is a neural network that is trained on our labeled data. 
Close-to-shore tracking improves with an accurate map and is demonstrated on a recently gathered real-world dataset. The dataset contains multiple sequences of a kayak and a day cruiser moving close to the dock, in a collision path with an autonomous ferry prototype. 
\end{abstract}


\glsresetall

\section{Introduction}

\IEEEPARstart{O}{bject} detection and tracking is an essential part of the situational awareness system for an autonomous vehicle. For example, as an autonomous ferry navigates its environment, other vessels, such as kayaks and boats, can come into collision course with the ferry. Their paths will then need to be predicted for the ferry to move accordingly to avoid collision and follow the accepted movement patterns at sea. These near-shore relatively smaller vessels should therefore be detected and tracked. 

\Gls{lidar} is one of the most common sensors for autonomous vehicles, especially for autonomous ferries. This is likely due to its accurate and precise measurements of the environment. However, cameras often have denser measurements with rich information, while the field of view is often smaller, and the data is more complex to utilize. For example, the image can be used to detect vessels at relatively far distances and the \gls{lidar} can be used to get accurate position estimates for tracking the vessel near shore~\cite{Hilmarsen2025}. 

\begin{figure}
    \centering
    \includegraphics[width=0.7\linewidth]{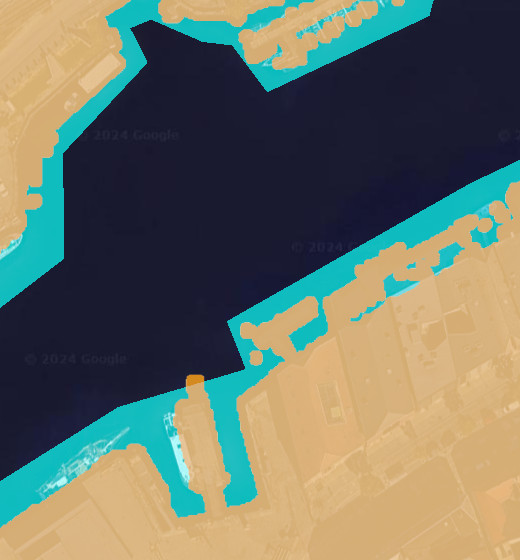}
    \caption{Overlain in cyan is an example of land masking. Overlain in orange is a visual and \gls{lidar}-based map, details are in \cref{sec:mapping}. The orange map is more detailed, without potentially moving objects, and gives earlier target detections than the cyan map. A photo of a relevant docking area is in the background. Base image courtesy of Google Maps: Imagery @2024 Airbus, CNES / Airbus, Maxar Technologies, Map data @2024. }
    \label{fig:intro_land_masking}
\end{figure}

\begin{figure}
    \centering
    \includegraphics[width=0.8\linewidth]{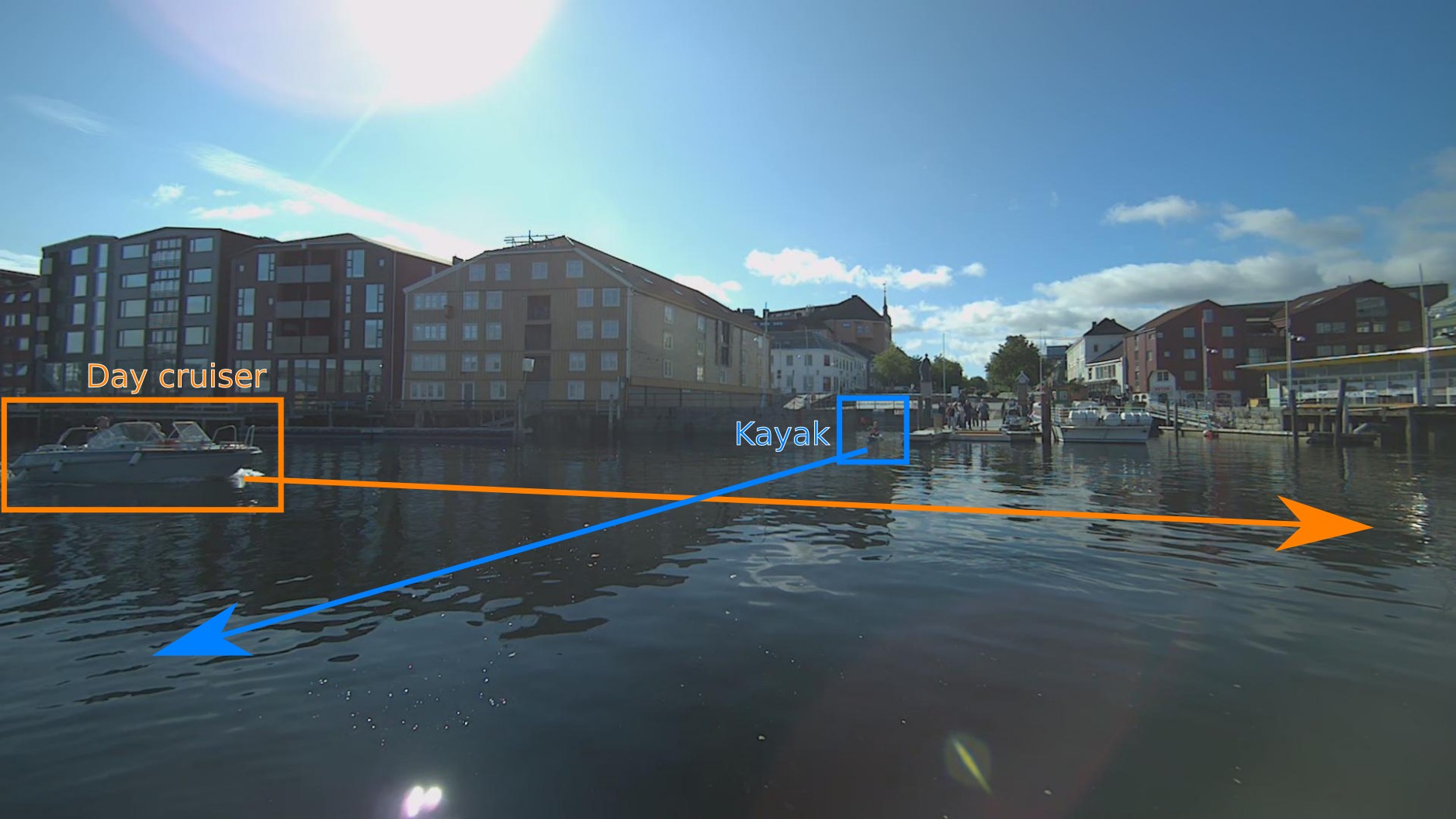}
    \caption{The kayak is undocking and moving into the canal. The approximate path it takes is marked in blue, and the approximate path for the day cruiser is marked in orange. A precise map enables earlier tracking of the kayak compared to an imprecise map, where the kayak can be differentiated from the dock. }
    \label{fig:intro_kayak_near_dock}
\end{figure}

As can be seen in \cref{tab:related_work_topics}, many methods use \gls{lidar} for \gls{mos} or \gls{datmo}. However, we suggest distinguishing between static and potentially dynamic objects during mapping. A mapping method that only does \gls{mos}, or \gls{datmo}, maps objects that are static during the mapping but are potentially moving objects, such as docked vessels. This is especially relevant in harbor areas, where a lot of vessels are docked, and a lot of traffic occurs. The map is then wrong if the docked vessel moves. Even worse, an autonomous ferry is approaching the dock and it does not detect that an earlier static vessel now moves into a potential collision area since the target vessel has not moved sufficiently from its docking position. 

Target tracking in the maritime domain typically employs land masking, e.g., see \cref{fig:intro_land_masking}. Land masking is where the areas on land and a bit into the water are marked. This is so that the land, and sometimes also static vessels, can be filtered out of the sensor data, resulting in only data from vessels that one would like to track. This has been done for tracking in \gls{radar} data in \cite{Wilthil2017, Wilthil2018Track} and in \gls{lidar} data in \cite{Helgesen2022}. However, it is often not precise enough to also detect vessels that are close to the quay. For example, kayaks are relatively nimble, meaning that a kayak might suddenly get into a collision path with the ferry when the ferry itself is close to the quay, see \cref{fig:intro_kayak_near_dock}. To avoid colliding with these kayaks, it is important to detect and track them as early as possible. 

To detect objects early, we suggest using a precise map that makes it easier to find what measurements do not correspond to static land and should be tracked for collision avoidance. In this paper, we use the appearance of the objects in images to evaluate this, with the use of a deep learning-based instance segmentation technique. 
The contributions of this paper are as follows:
\begin{itemize}
    \item We present a method for maritime \gls{lidar} and camera mapping that does not map potentially moving objects.
    \item We demonstrate the improved performance of close-to-shore target tracking with the use of a precise map, where targets are detected early, compared to less precise maps. 
    \item The method is demonstrated on real-world data. 
\end{itemize}

\section{Related work}

\begin{table*}[hbtp]
    \caption{This table shows the different topics for related work. ``Free space'' means the method explicitly estimates free space. With ``ENC'' we include the use of OpenStreetMap in the automotive domain. }
    \label{tab:related_work_topics}
    \centering
    \resizebox{0.99\linewidth}{!}{%
    \begin{tabular}{c|c|c|c|c|c|c|c|c|c|c|c}
         & Camera & LiDAR & RADAR & SLAM & \makecell{MOS / \\DATMO} & Tracking & \makecell{Free \\ space} & \makecell{Deep \\ learning} & \makecell{Change \\ detection} & ENC & \makecell{Maritime \\ domain}\\
         \hline 
        \cite{Chen2022a}            &   & X &   & X & X & X & X &   &   &   &   \\
        ERASOR~\cite{Lim2021}       &   & X &   & X & X & X & X &   &   &   &   \\
        Removert~\cite{Kim2020}     &   & X &   & X & X & X &   &   &   &   &   \\
        OctoMaps~\cite{Arora2021}   &   & X &   & X & X & X & X &   &   &   &   \\
        \cite{Pfreundschuh2021}     &   & X &   & X & X & X & X &   &   &   &   \\
        \cite{Jinno2019}            &   & X &   &   &   &   & X &   & X &   &   \\
        \cite{Ding2018}             &   & X &   & X &   &   & X &   & X &   &   \\
        \cite{Pomerleau2014}        &   & X &   & X & X & X & X &   & X &   &   \\
        \cite{Mersch2023}           &   & X &   & X & X & X &   & X &   &   &   \\
        \cite{Hosseinyalamdary2015} &   & X &   &   &   & X & X &   &   & X &   \\
        \cite{Wang2003}             &   & X &   & X & X & X & X &   &   &   &   \\
        \cite{Miyasaka2009}         &   & X &   & X & X & X & X &   &   &   &   \\
        \cite{Thompson2019}         &   & X &   &   &   &   &   &   &   &   & X \\
        \cite{Xie2023}              &   & X &   &   & X &   &   &   &   &   &   \\
        \cite{Chen2021a}            &   & X &   &   & X &   &   &   &   &   &   \\
        \cite{Postica2016}          & X & X &   &   & X &   & X &   &   &   &   \\
        \cite{Yan2014}              & X & X &   &   & X & X & X &   &   &   &   \\
        OGM-HMM~\cite{Sun2024}      &   &   & X &   & X & X &   &   &   & X & X \\
        \cite{Dewan2016a}           & X &   &   &   & X & X &   &   &   &   &   \\
        \cite{Dewan2016}            & X &   &   &   & X & X &   &   &   &   &   \\
        \cite{Lenz2011}             & X &   &   &   & X & X &   &   &   &   &   \\
        \cite{Lee2010}              &   & X &   &   & X & X &   &   &   &   & X \\
        \cite{Nuss2018}             &   & X &   & X & X & X & X &   &   &   &   \\
        \cite{Negre2014}            &   & X &   & X & X & X & X &   &   &   &   \\
        \cite{Tanzmeister2014}      &   & X &   & X & X & X & X &   &   &   &   \\
        \cite{Schreiber2021}        &   & X &   & X & X & X & X &   &   &   &   \\
        \cite{Vatavu2020}           & X & X &   & X & X & X & X &   &   &   &   \\
        \cite{Hilmarsen2025}            & X & X &   &   & X & X &   & X &   &   & X \\
        \cite{Bibby2010}            &   &   & X & X & X & X & X &   &   & X & X \\
        \cite{Vu2011}               &   & X & X & X & X & X & X &   &   &   &   \\
        \cite{Gies2018}             & X & X & X & X & X & X & X &   &   & X &   \\
        \cite{Pieper2024}           &   & X &   &   &   &   &   &   &   & X & X \\
        \cite{Obradovic2024}        & X & X &   &   & X & X &   & X &   &   & X \\
        \cite{Yao2023}              &   & X &   &   & X & X &   &   &   &   & X \\
        \cite{Lin2022}              &   & X &   &   & X & X &   & X &   &   & X \\
        \cite{Chi2024}              & X & X &   & X & X & X &   & X &   &   &   \\
        \textbf{Ours}               & \textbf{X} & \textbf{X} &   &   &   & \textbf{X} &   & \textbf{X} &   & \textbf{X} & \textbf{X}
    \end{tabular}
    }
\end{table*}

Many of the methods in \cref{tab:related_work_topics} rely on occupancy grid maps, which explicitly model free space. However, in the maritime domain, this approach poses challenges because the \gls{lidar} detects very few points on water. Interpreting the absence of measurements as the presence of water may not be reliable. 
When generating a 2D map, a 3D \gls{lidar} cannot accurately measure free space in 2D unless it explicitly detects the points where the laser hits the water. Additionally, many methods rely on ground segmentation, which becomes problematic when water is not detected.

However, some methods have been applied in the maritime domain. \cite{Pieper2024} used \gls{enc} and mapped the deviations. This was for autonomous surface vessels to move safely without colliding into static structures that are not in the \gls{enc}. 
\cite{Yao2023} did near-shore multi-object tracking and static mapping using \gls{lidar}. They could then not detect static vessels. Furthermore, they argue against using cameras since cameras are sensitive to rain, fog, adverse weather conditions, and occlusions. This is also a reason why we split the mapping and tracking tasks so that the tracking does not require the camera when the weather is bad. 

Our earlier work \cite{Hilmarsen2025} did near-shore target tracking. However, it only detected moving objects unless they were in the image. That reduces the field of view of the tracking. This might work when moving fast forward like in the automotive domain. However, for the maritime domain, other vessels are not limited to the same location and movement patterns as on roads and we argue that it is beneficial to use the field of view of the \gls{lidar}. 
  
The most relevant work to ours is \cite{Chi2024}, where the authors performed \gls{slam} in the automotive domain. They utilized car detections in both images and \gls{lidar} point clouds to filter out dynamic objects, enabling the mapping of only the static environment. They had available a mature Yolov7 detector~\cite{Wang2023} for the automotive domain, which we do not have for the maritime domain. They were also limited to the field of view of the camera.

\section{Mapping}

\label{sec:mapping}

The mapping process is done offline and combines context-aware segmentation from images with precise point measurements from a \gls{lidar}. It uses a neural network-based learning method for potentially moving object segmentation in the images, which is detailed in \cref{sec:potentially_moving_object_segmentation}. 

This section describes how these two sensor modalities are combined to get a precise 2D map: First, an \gls{enc} is used for initialization. Then, \gls{lidar} measurements are accumulated. Only regions in an image that are not regarded as a vessel are stored in the map. Lastly, the map is post-processed to remove inconsistencies.

\subsection{Point Selection}

As new measurements are received from the \gls{lidar}, it is decided whether each point comes from a vessel or not. This is done using the before-mentioned image segmentation method. However, each point measurement has to be projected into the camera first. 

First, each point is transformed to the reference frame of the camera:
\begin{equation}
    \widetilde{\mathbf{X}}^{\mathrm{cam}} = \mathbf{H}^{\mathrm{cam}}_{\mathrm{LiDAR}} \widetilde{\mathbf{X}}^{\mathrm{LiDAR}}, 
    \label{eq:mapping_transformation}
\end{equation}
where $\widetilde{\mathbf{X}}^{(\cdot)}$ is the 3D homogeneous coordinate of a point in the $(\cdot)$ reference frame, and $\mathbf{H}^{\mathrm{cam}}_{\mathrm{LiDAR}}$ is the homogeneous transformation matrix between the reference frames. The transformation matrix is assumed known from calibration. 

Then, each point is projected into the camera using the standard projection formula~\cite[p. 472]{forstner2016photogrammetric}:
\begin{equation}
    \begin{bmatrix}
        Z \cdot x \\
        Z \cdot y \\
        Z
    \end{bmatrix} = \mathbf{K} \mathbf{X}^{\mathrm{cam}} = \begin{bmatrix}
        f_x & 0 & c_x \\
        0 & f_y & c_y \\
        0 & 0 & 1
    \end{bmatrix} \begin{bmatrix}
        X \\
        Y \\
        Z
    \end{bmatrix}, 
    \label{eq:mapping_projection}
\end{equation}
where $X$, $Y$ and $Z$ are the real-world coordinates of the point in the camera reference frame with unit meters and $x$ and $y$ are the image coordinates of the projected point into the image with unit pixels. The matrix $\mathbf{K}$ is the camera intrinsic matrix and is also assumed known from calibration. 

After points are projected into the image, all points that are not within the field of view of the camera are removed since we cannot know if these points belong to a vessel or not. The image segmentation method gives a segmentation mask for each vessel detected in the image. We call these vessels ``potentially moving objects''. The points that fall within such a mask are stored as detections. The points that fall within the image, but are not within a detection mask, are stored as candidates for mapping.

\subsection{Point Accumulation}

An \gls{enc} is used for initialization of the map. These are pre-existing maps that are available that should include many of the detailed structures in regions of maritime traffic. However, they are not always as detailed as required for accurate target tracking near shore. The \gls{enc} enables the method to cover the key areas for mapping without requiring a detailed view of that area, unlike \gls{lidar} and camera-based approaches. We then create a grid out of the \gls{enc} which describes what level of details we want to store and what level of detailed view is required when mapping. Having smaller grid cells allows for finer details, but also requires the \gls{lidar} and camera data to measure enough details from all regions. 
The grid is in 2D. Each 3D point is transformed to the world frame from the \gls{lidar} frame using a similar transformation as in \cref{eq:mapping_transformation}. The transformation matrix is found using the \gls{ins} of the ego-vessel. Each point is then projected down to the horizontal plane and is reduced to being in one of the grid cells of the map. 

As new \gls{lidar} measurements are received, and each point is classified as unknown, a vessel detection or potential static structure, points are accumulated in a sliding window fashion. This is because a single image is not always perfectly segmented. Sometimes a vessel can be missed and sometimes land can be mistaken as a vessel. However, over time we can be more and more certain about how we segment the world. 
Over this sliding window, certain conditions need to be satisfied for a grid cell to become regarded as a static structure in the map: 
Firstly, the cell cannot be farther away than a specific distance threshold, this is a tunable parameter. This is because the vessel detections get worse at farther distances as well as the \gls{lidar} measurements being very sparse at great distances. 
Secondly, the cell must not be classified as a vessel at any point during the sliding window. If a cell is classified as a vessel, we must wait to determine whether sufficient evidence later supports its classification as land.
Lastly, the cell needs to be measured in at least a specific percentage of the frames within the sliding window, this is a tunable parameter. This ensures that sparse clutter from waves is not stored in the map, while allowing regions that are so far away that the \gls{lidar} does not consistently give measurements in that cell.

\subsection{Post-processing}

After the measurements have been accumulated in the grid and the sliding window has passed, each cell is reduced to being a binary grid of land or not land. The grid is then processed to remove inconsistencies. For example, there might be a single cell that was not stored as a static structure, while all its neighboring cells were. This is likely then also a static structure. The opposite is also possible, where a bird or clutter in the \gls{lidar} data has given rise to a single cell that is mistakenly regarded as a static structure surrounded by what is otherwise water. To address these issues we use a series of morphological operations: dilation and erosion. The finished map can be used in tracking as described next.

\section{Tracking}

After the map has been created, it can be used for removing \gls{lidar} points belonging to static structures in the environment. This is done in a similar way as during mapping, by transforming the points to the 2D world frame and removing points that fall within a cell in the map that corresponds to a static structure. The 3D points in the \gls{lidar} point cloud that do not fall within such a cell can still be kept in 3D. However, we opt for 2D tracking in this paper as it is sufficient for demonstration. 

There are reasons to only use \gls{lidar} for tracking and not the camera. It would be possible to use the image boat detections for tracking as well, but relying on it would enforce the field of view of the camera. The field of view can therefore be bigger when using only the \gls{lidar}, unless the camera has really wide field of view.

\subsection{Detection}

For simplicity, we choose to track 2D boat detections. While 3D \gls{eot} might give more accurate tracking, 2D point tracking is sufficient to demonstrate the improved tracking with a map from our mapping method compared to less detailed maps. 

The boat detections are based on \gls{dbscan}~\cite{ester1996density} run on the point cloud projected to 2D that is not considered a part of the static structures in the map. This is a clustering method that groups the points into clusters based on the distance between groups of points and the density inside the groups. Each cluster of points is then reduced to being a single-point measurement that can be processed in a point-target tracker.

\subsection{VJIPDA Tracker}

The point-target tracker utilized here is the \gls{vjipda} tracker~\cite{Brekke2021}. We choose the \gls{vjipda} as its visibility state can improve the tracking during occlusions that are common in near-shore environments.

\section{Potentially Moving Object Segmentation}
\label{sec:potentially_moving_object_segmentation}

\begin{figure}
    \centering
    \includegraphics[width=0.7\linewidth]{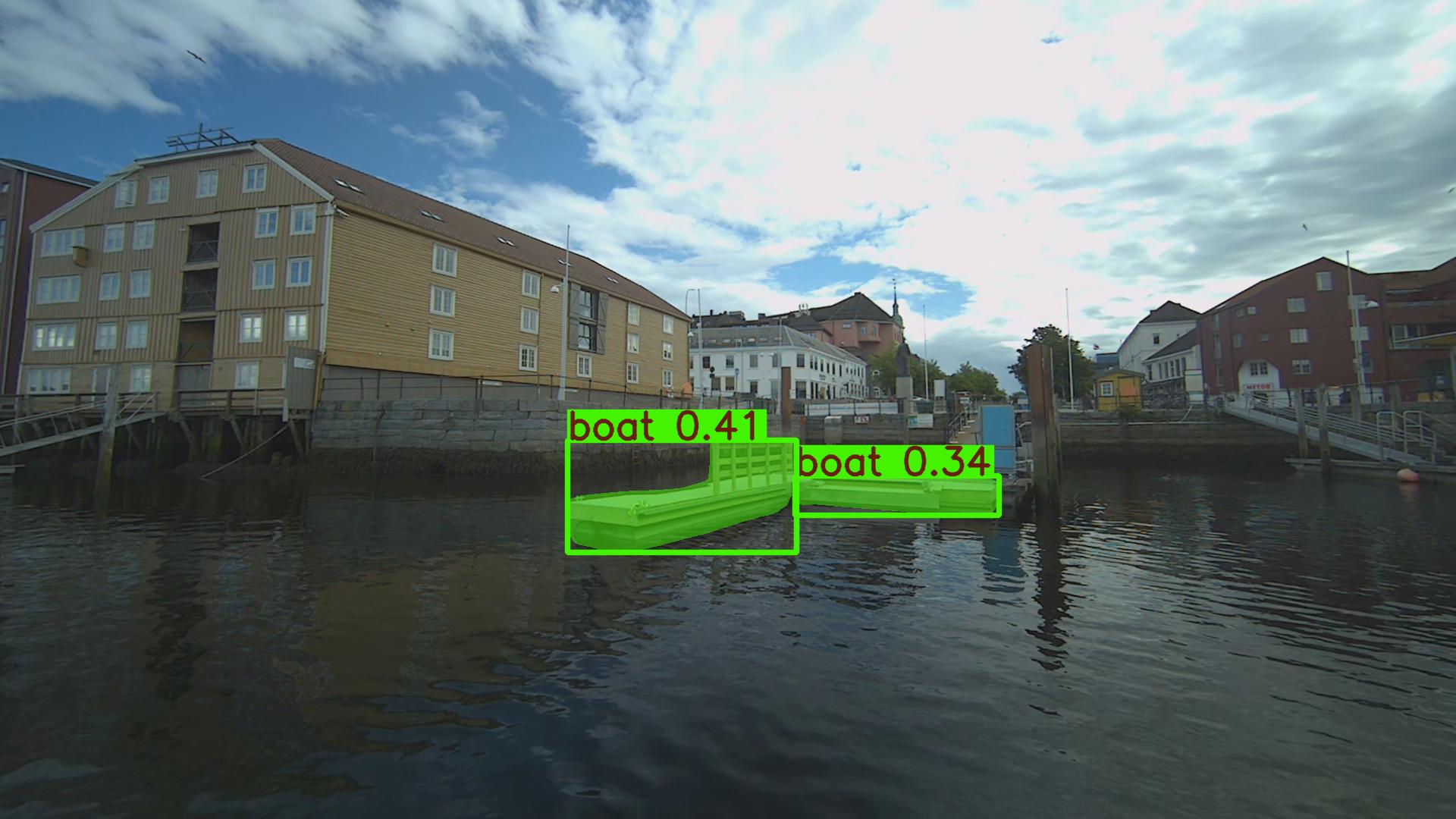}
    \caption{The static floating dock is mistakenly detected as two boats. This needs to be handled to map the dock. }
    \label{fig:seg_mistake_dock}
\end{figure}

By ``potentially moving object'' we include static boats as well as moving boats. To detect both of these, we utilize the camera and an instance segmentation method for the images. The instance segmentation method used in this paper is Yolov8~\cite{Jocher_YOLO_by_Ultralytics_2023}. It is relatively fast and gives good enough detections for the most part. However, the masks are not always how we would like them. \Cref{fig:seg_mistake_dock} shows the challenge where the floating dock is detected as a boat. This happened consistently when using the pre-trained model that was trained on the COCO dataset~\cite{Lin2014}, even with detection scores of up to $80\%$. To alleviate this challenge, we trained the network on more relevant data. 

There are several datasets available for training in addition to COCO. Firstly, the MariBoats dataset~\cite{Sun2023} consists of $6.2 \mathrm{k}$ images with instance segmentation of the class ``ship''. The data was gathered by searching online for images. The images are not very representative of the autonomous ferry scenario, with many big and industrial ships and plenty of text on the images. It is unlikely that such a dataset will clearly differentiate between the floating dock used here and a raft or something similar. We tried training on this data, but it did not improve the situation in \cref{fig:seg_mistake_dock}. Secondly, ABOships~\cite{Iancu2021} has boat detections from the autonomous ferry perspective. However, they have not created instance segmentation annotations, only bounding box detections. 

\begin{figure}
    \centering
    \subfloat[Annotation of boats and no annotation on the floating dock. ]{%
      \includegraphics[width=0.7\linewidth]{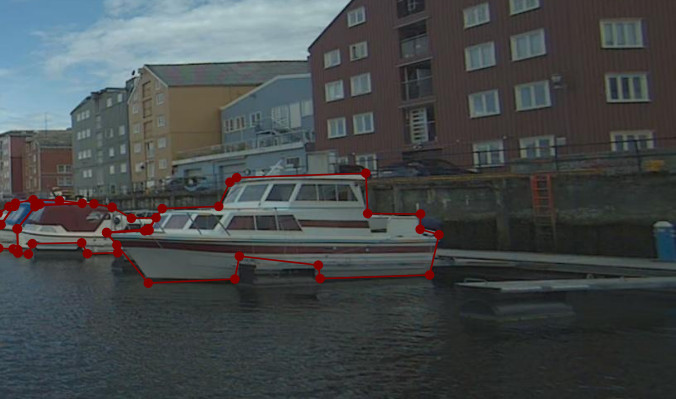}
    }
    
    \subfloat[Purposefully not labeled dock. ]{%
      \includegraphics[width=0.7\linewidth]{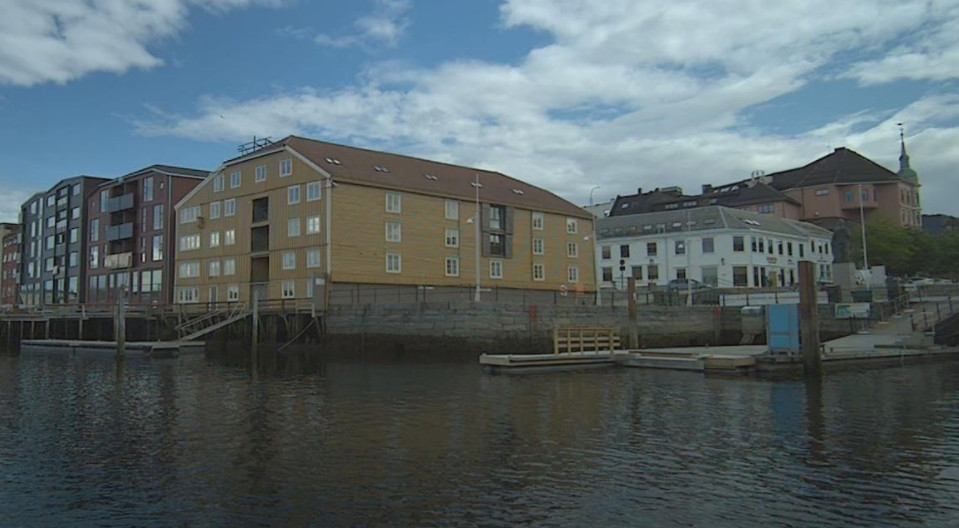}
    }
    
    \caption{Examples of annotations that differentiate boats from floating docks. This labeling is used to train the segmentation method. }
    \label{fig:seg_annotations}
\end{figure}

It is not obvious that the floating dock is not a boat. Even a human could think it is a boat if the image is not taken very close to the dock. Therefore, for an instance segmentation method to work well, we choose to create our own annotations on our own dataset to teach the network that it is a dock and not a boat. 

The way we choose to train the network is by training in multiple steps. Firstly we change the COCO dataset~\cite{Lin2014} to use all images with boat annotations, which is $3146$ images, and $5\%$ of the other images, about $6 \mathrm{k}$ images. The whole network is then trained on this. Secondly, we freeze the first $5$ layers of the network and train the rest on the ABOships dataset~\cite{Iancu2021}, with the whole detection boxes as masks. The idea here is that the finer segmentation can be learned by the last layers and that we want to utilize more data than only COCO. Lastly, the network is trained on our own data with the first $10$ layers frozen. The result is a method that manages most detections correctly with fewer false positive detections on the docks. 
This could in a sense be seen as implicit mapping or over-training. However, having a good enough detector is necessary when using images in some specific scenarios. Our dataset has $181$ images with instance segmentation annotations of boats. Examples are shown in \cref{fig:seg_annotations}. Note that the floating docks are not labeled as boats.

\section{Results}

\subsection{Dataset}

The data used in this paper has not yet been published, but we intend to publish it soon. Without going into all the details of the dataset, we can explain the most important aspects. The sensors used are \gls{ins}, \gls{lidar}, two stereo cameras and two \gls{gnss} recorders for reference tracks. The \gls{ins} uses \gls{rtk} \gls{gnss} to get accurate position and combines it with an \gls{imu} to get accurate orientation. The sensors have all been calibrated with respect to each other. 
The \gls{ins} runs at $100 \mathrm{Hz}$, the cameras at $30$ fps and the \gls{lidar} at $10 \mathrm{Hz}$. All sensors have been synchronized to \gls{gnss} time, although they are not triggered simultaneously.

\subsection{Implementation}

We will mention the parameters used in the mapping method. The sliding window is set to $5\mathrm{s}$, that is $50$ frames. The distance threshold for the mapping is $100\mathrm{m}$. The percentage of frames needed to be measured for a cell to be regarded as a static structure is $40\%$. The grid size is $0.5\mathrm{m}$. 

\begin{figure}
    \centering
    \includegraphics[width=0.9\linewidth]{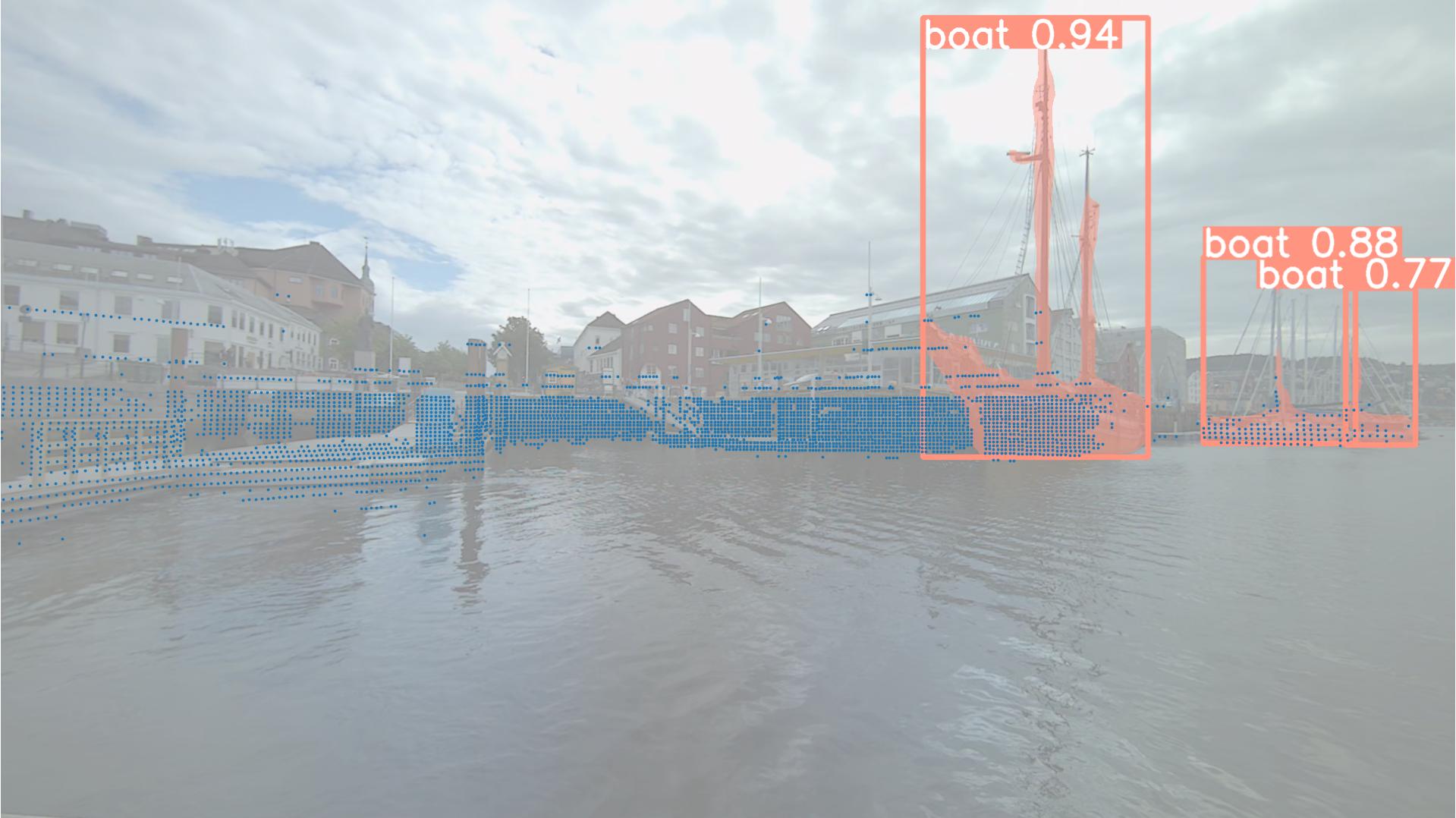}
    \caption{What a single frame can look like during mapping. The boat detections are in light orange, the image is brightened in the background and the projected \gls{lidar} points are in blue. This illustrates how the point selection is done, differentiating between static objects and vessels. }
    \label{fig:map_image_lidar_detections}
\end{figure}

An illustration of the \gls{lidar} point projection to the image and the image segmentation is shown in \cref{fig:map_image_lidar_detections}. We can see that some large ships are detected correctly with good masks and that the docking area is not detected as a vessel. 

The \gls{enc} data used was sourced from OpenStreetMap~\cite{OpenStreetMap} and obtained using the OSMnx tool~\cite{boeing2024modeling}. The polygons were found and reduced to only the interesting area.

\subsection{The Resulting Map}

\begin{figure}
    \centering
    \subfloat[All \gls{lidar} points have been accumulated into a map. Three boats have been circled in red to show that these have become a part of the map.\label{fig:map_lidar_accumulate_all}]{%
      \includegraphics[width=0.45\linewidth]{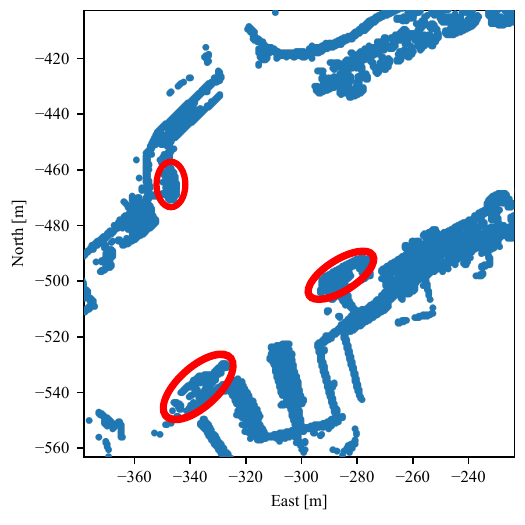}
    }\hfill
    \subfloat[The \gls{enc} from \href{https://www.openstreetmap.org/copyright}{OpenStreetMap} is shown in blue. The map created from accumulating segmented \gls{lidar} points is shown in orange.\label{fig:map_lidar_and_enc}]{%
      \includegraphics[width=0.45\linewidth]{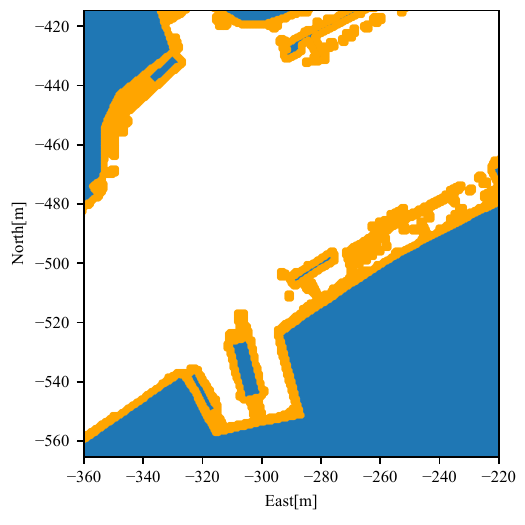}
    }
    
    \caption{Comparison of two different maps. Our mapping method improves the differentiation between land and vessels. }
\end{figure}

If the mapping is done naively, without considering potentially moving objects, one might get a map similar to \cref{fig:map_lidar_accumulate_all} where also the docked boats are mapped. As mentioned before, this might make the tracking algorithm miss these boats, which could potentially result in a collision with the ferry. The resulting map using our mapping method is shown in \cref{fig:map_lidar_and_enc}. There we see that the boats are not stored and the \gls{enc} fills a lot of the unknown space in the pure accumulation map. Post-processing the map is particularly useful for sections of the walkway near land that lacks \gls{lidar} points due to the sensor's sparsity at that distance, leading to gaps in the data.

\subsection{Tracking Using Different Maps}

The main differences between not using a map, using less accurate maps and using accurate maps for tracking are the number of false tracks and the time of track initialization, not necessarily the precision of each track. Therefore, we will compare the use of different maps and show both qualitatively and quantitatively the difference. 

An example of the tracking results is shown in \cref{fig:track_result}. The kayak that undocks from the quay has been tracked while it is close to the quay. A day cruiser comes in front, occluding the kayak, but the kayak is tracked again afterward because of the accurate enough predictions of its movements. 

\begin{figure*}
    \centering
    \subfloat[Tracking without a map. The list of tracks is too long for the image. The number of tracks became $33$.\label{fig:result_tracking_without_map}]{%
      \includegraphics[width=0.32\linewidth]{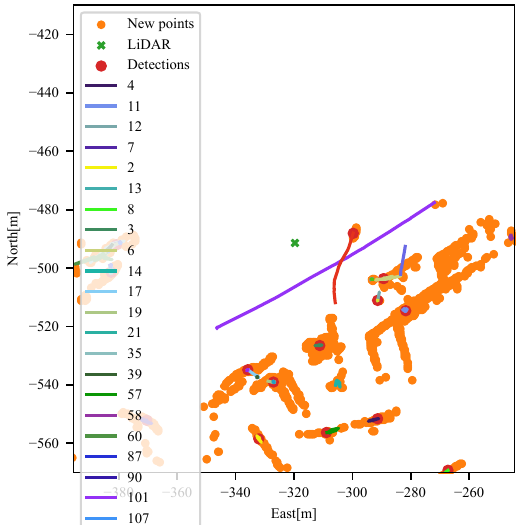}
    }\hfill
    \subfloat[Tracking using only the \gls{enc} as map. Gave rise to $22$ tracks.\label{fig:result_tracking_with_only_enc}]{%
      \includegraphics[width=0.32\linewidth]{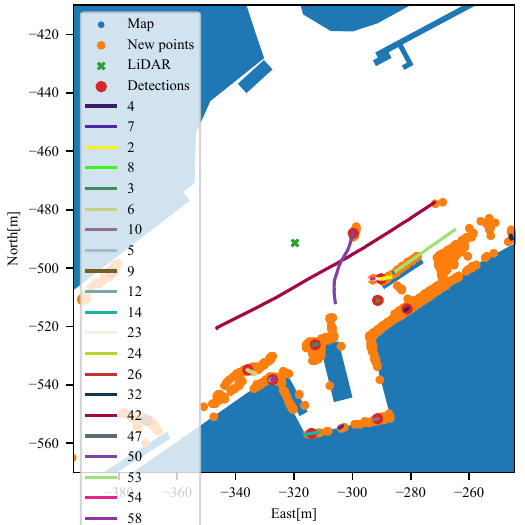}
    }\hfill
    \subfloat[Tracking in a map with big margins.\label{fig:result_tracking_with_dilated_map}]{%
      \includegraphics[width=0.32\linewidth]{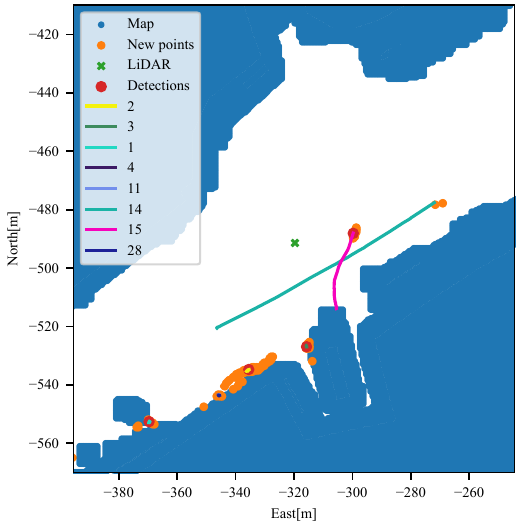}
    }
    
    \caption{What the measurements, detections and tracks look like when using imperfect maps. }
    \label{fig:result_different_maps}
\end{figure*}

\begin{figure}
    \centering
    \includegraphics[width=0.7\linewidth]{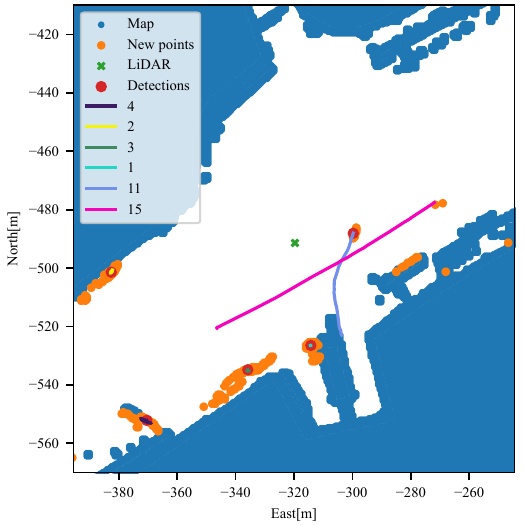}
    \caption{Tracking with a good map, where all vessels close to the ferry have been detected and tracked. The tracking happens earlier than in the worse maps in \cref{fig:result_different_maps}. }
    \label{fig:track_result}
\end{figure}

Tracking all detections without a map can cause a lot of false tracks, see \cref{fig:result_tracking_without_map}. The same detection and tracking as earlier is used, just without any map. It results in $33$ tracks, while the correct number is $6$. While it is possible to do better boat detections in the \gls{lidar} point clouds, this shows that it is beneficial to remove the points that are known to be land. We also see that the kayak that is close to the dock is first detected when it comes far enough from the dock to not be clustered with it. The corresponding track in \cref{fig:track_result} starts when the kayak is right next to the dock, see id $11$. The day cruiser that crosses the path of the kayak is detected and tracked similarly in the two cases. 

Using only \gls{enc} as a map also gives rise to false tracks. This is shown in \cref{fig:result_tracking_with_only_enc}. The reason for this is the structures that are not in the map but are still detected.  Other than the number of false tracks, the results are similar to the ones with no map.

We have approximated a map based on land masking in \cref{fig:result_tracking_with_dilated_map}, with added tracking results, by using our estimated map in \cref{fig:map_lidar_and_enc} and adding a margin of $2 \mathrm{m}$. 
This is still more detailed than the land masking in \cref{fig:intro_land_masking}. Also, if the land masking had been made using the photo in \cref{fig:intro_land_masking}, and not with the details from the mapping done in this paper, then many regions would not have been masked correctly since the photo is not up to date. We can see that the points from the land are removed. However, points on boats have also been removed. Some boats are not detected at all, and the kayak is not detected before coming sufficiently far from the dock.

\subsection{Tracking in More Scenarios}

\begin{figure*}
    \centering

    \subfloat[Multi target cross. ]{%
      \includegraphics[width=0.19\linewidth]{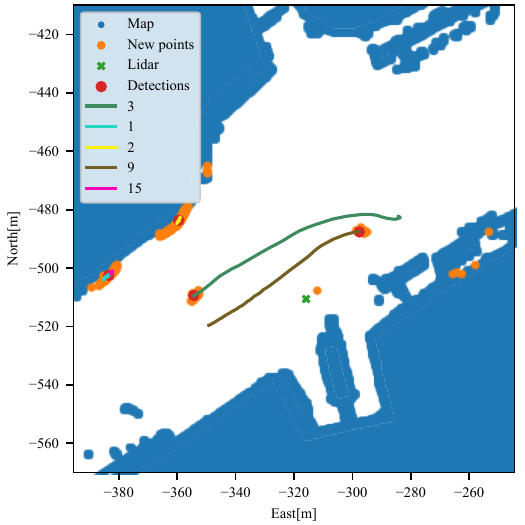}
    }\hfill
    \subfloat[Multi target day cruiser undock. ]{%
      \includegraphics[width=0.19\linewidth]{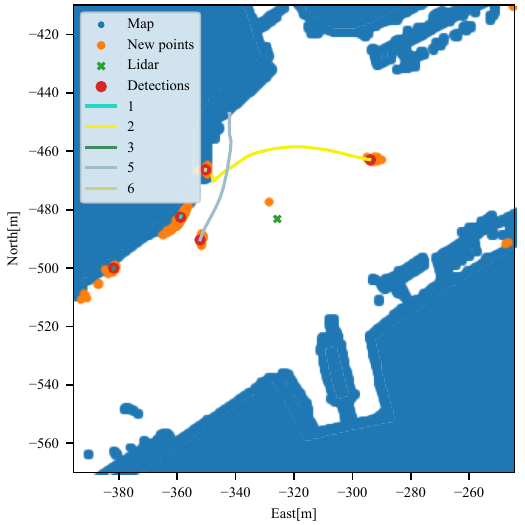}
    }\hfill
    \subfloat[Multi target kayak undock 1. ]{%
      \includegraphics[width=0.19\linewidth]{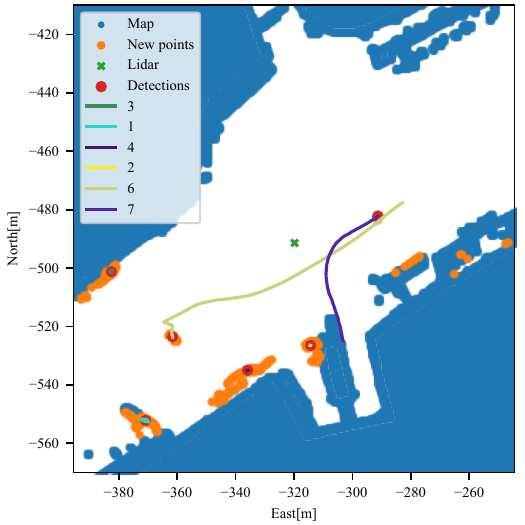}
    }\hfill
    \subfloat[Multi target kayak undock 2. ]{%
      \includegraphics[width=0.19\linewidth]{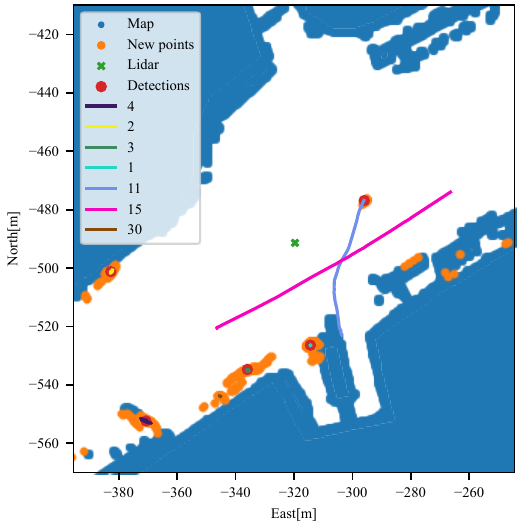}
    }\hfill
    \subfloat[Multi target maneuver 1. ]{%
      \includegraphics[width=0.19\linewidth]{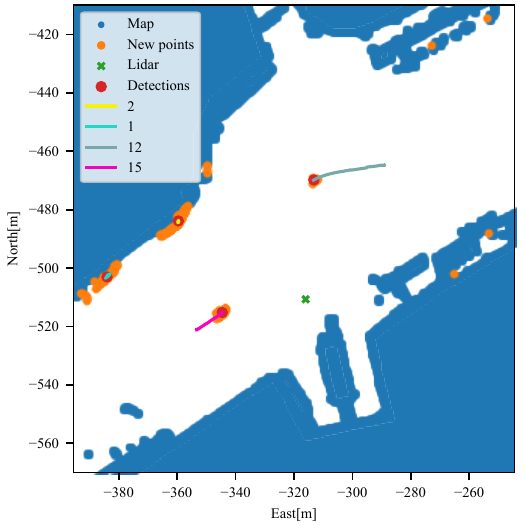}
    }
    
    \subfloat[Multi target maneuver 2. ]{%
      \includegraphics[width=0.19\linewidth]{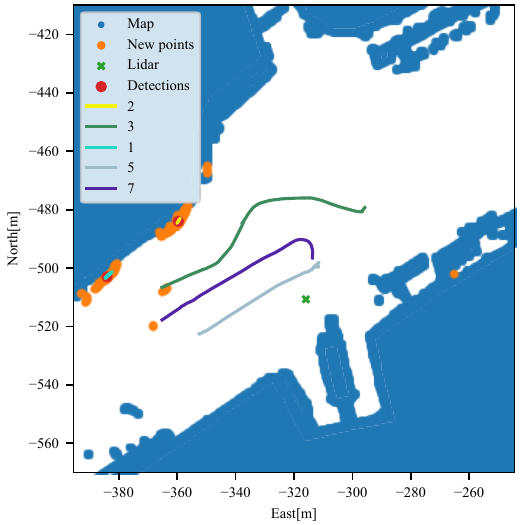}
    }\hfill
    \subfloat[Multi target pass 1. ]{%
      \includegraphics[width=0.19\linewidth]{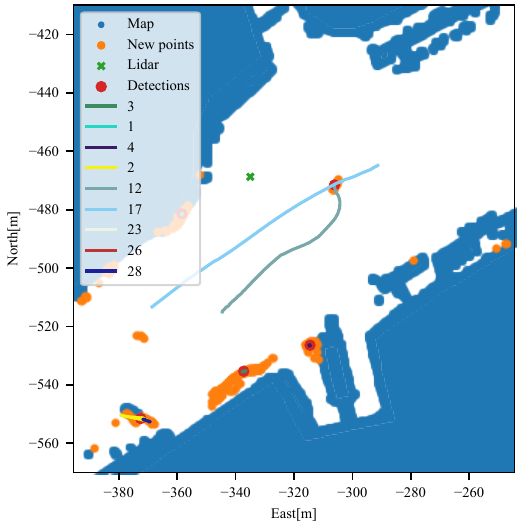}
    }\hfill
    \subfloat[Multi target pass 2. ]{%
      \includegraphics[width=0.19\linewidth]{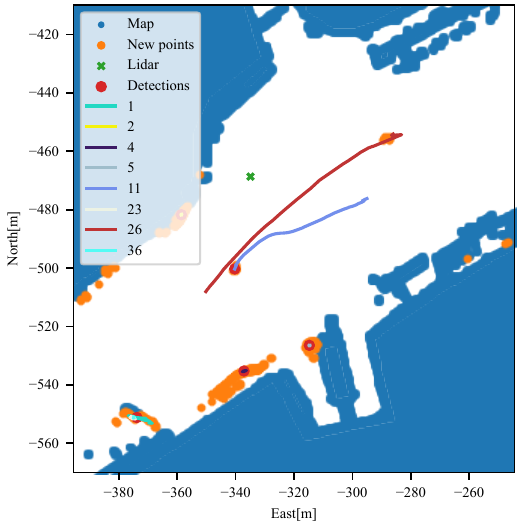}
    }\hfill
    \subfloat[Multi target pass 3. ]{%
      \includegraphics[width=0.19\linewidth]{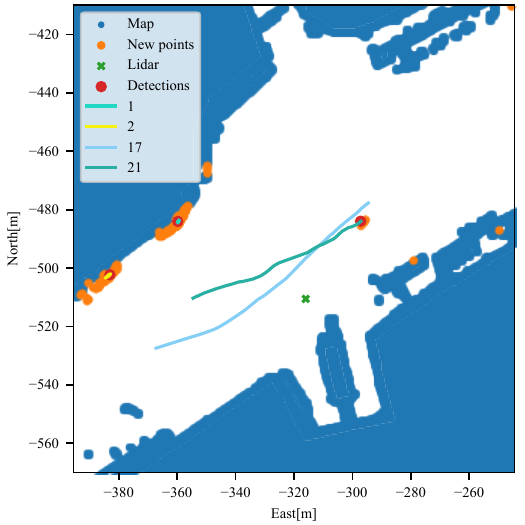}
    }\hfill
    \subfloat[Multi target pass 4. ]{%
      \includegraphics[width=0.19\linewidth]{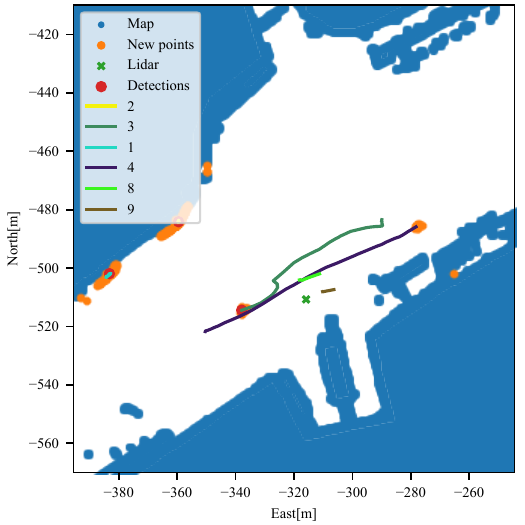}
    }
    
    \subfloat[Multi target pass 5. ]{%
      \includegraphics[width=0.19\linewidth]{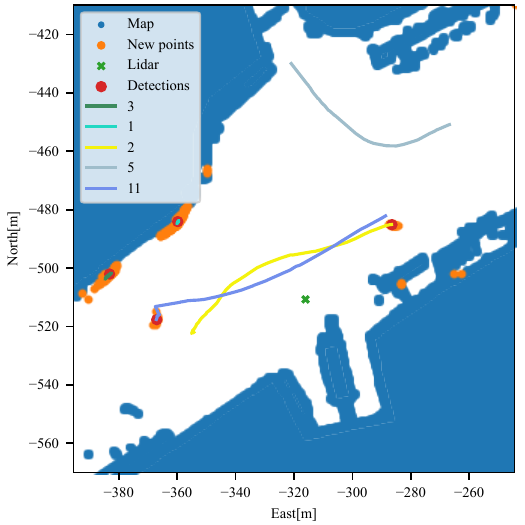}
    }\hfill
    \subfloat[Single target maneuver 1. ]{%
      \includegraphics[width=0.19\linewidth]{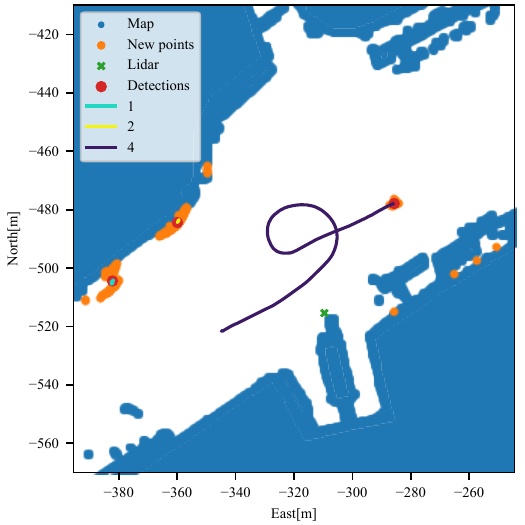}
    }\hfill
    \subfloat[Single target maneuver 2. ]{%
      \includegraphics[width=0.19\linewidth]{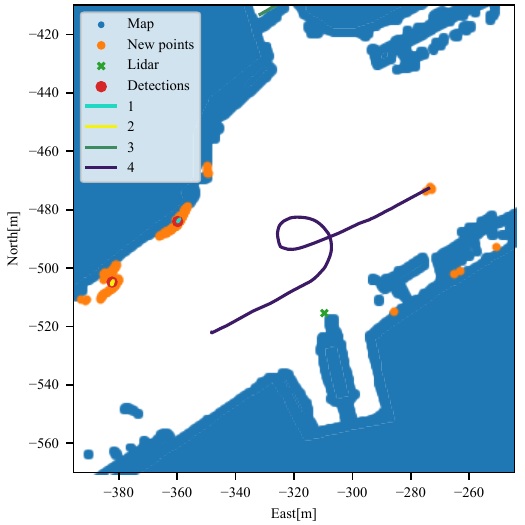}
    }\hfill
    \subfloat[Single target pass north. ]{%
      \includegraphics[width=0.19\linewidth]{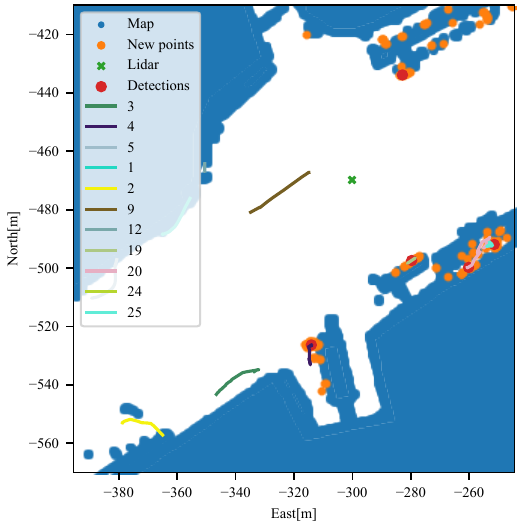}
    }\hfill
    \subfloat[Single target pass south. ]{%
      \includegraphics[width=0.19\linewidth]{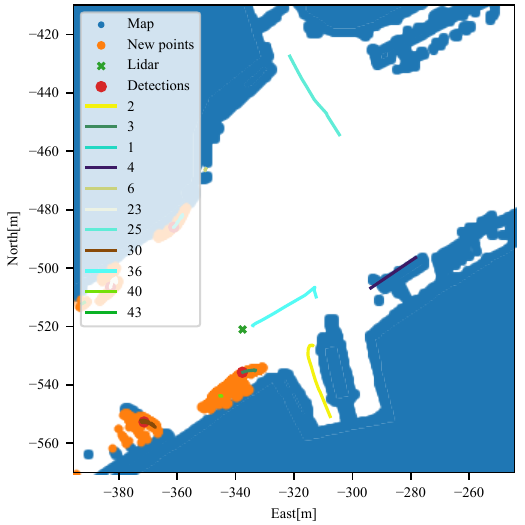}
    }
    
    \subfloat[Single target undock south. ]{%
      \includegraphics[width=0.19\linewidth]{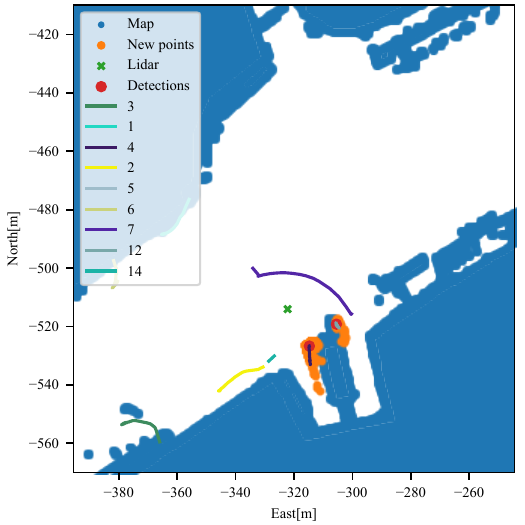}
    }\hfill
    \subfloat[Single target undock still 1. ]{%
      \includegraphics[width=0.19\linewidth]{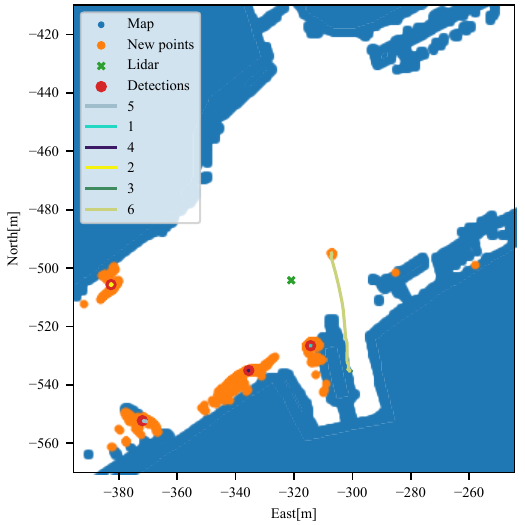}
    }\hfill
    \subfloat[Single target undock still 2. ]{%
      \includegraphics[width=0.19\linewidth]{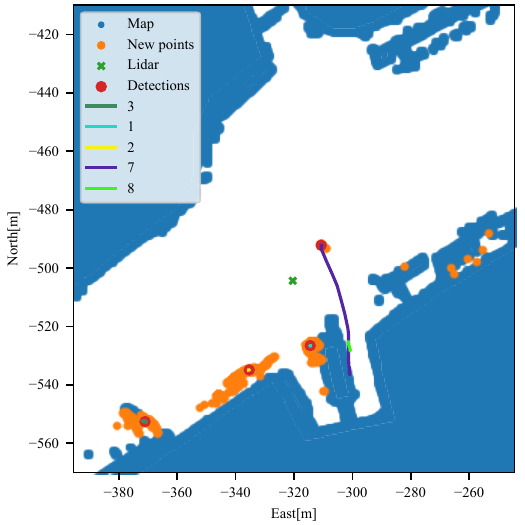}
    }\hfill
    \subfloat[Single target west 1. ]{%
      \includegraphics[width=0.19\linewidth]{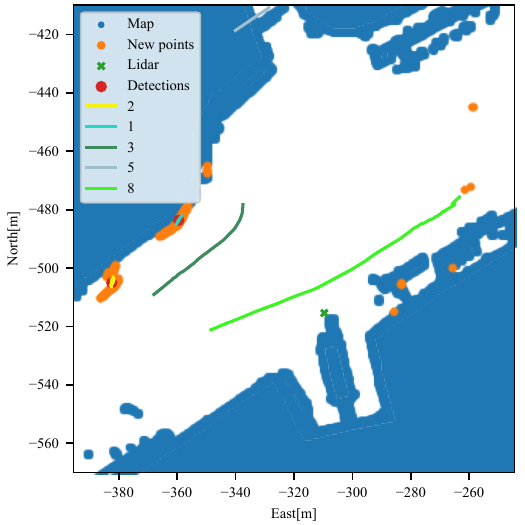}
    }\hfill
    \subfloat[Single target west 2. ]{%
      \includegraphics[width=0.19\linewidth]{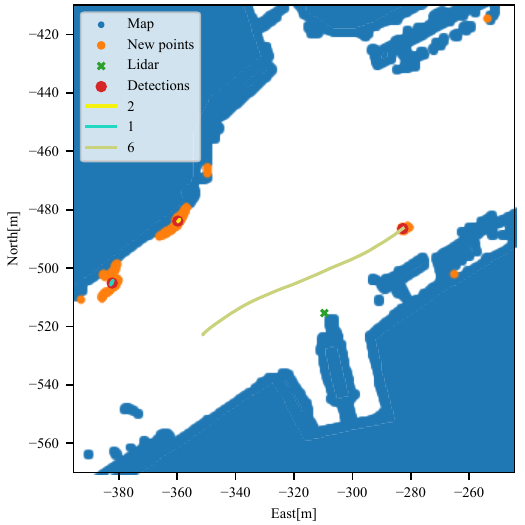}
    }
    
    \caption{All $20$ sequences we have gathered are plotted with tracks from the method in this paper. }
    \label{fig:result_all_sequences}
\end{figure*}

The tracking is performed on all our $20$ sequences, and the final tracks are shown in \cref{fig:result_all_sequences}. Similarly to the tracking in maps with fewer details, some of these sequences also show false tracks. This is due to certain regions not being mapped well. We used a specific mapping sequence to do the mapping, and not all regions were then seen in detail with the camera. A solution to this could be more detailed mapping, mapping on the tracking sequences as well, or an online mapping and tracking approach. Furthermore, one of the three maneuvering sequences shows what can happen when the constant velocity model is not good enough, where track is lost and re-created as the target does a quick maneuver.

\section{Conclusion}

In this paper, we proposed a method of computer vision-aided \gls{lidar} mapping to create detailed maps for reliable vessel detection and tracking.
Our approach demonstrated that integrating precise \gls{lidar} data with a trained image segmentation method enabled detection of potentially moving objects.
By using the camera solely for offline mapping, we ensured full field of view of the \gls{lidar} sensor during tracking. The results confirmed the method's effectiveness, even in detecting close-to-dock vessels before they entered potential collision zones.
The main limitations of this method is that it requires a decent vessel detector and that the whole region is mapped sufficiently. However, detectors are constantly being improved and some maritime systems repeatedly operate in specific regions. This makes the method highly relevant for systems such as autonomous ferries.

\section{Acknowledgements}
We use some map data copyrighted OpenStreetMap contributors and available from \url{https://www.openstreetmap.org}.

\bibliographystyle{IEEEtran}
\bibliography{main.bib}

\begin{thebibliography}{10}
\def\url#1{}
\csname url@rmstyle\endcsname
\providecommand{\newblock}{\relax}
\providecommand{\bibinfo}[2]{#2}
\providecommand\BIBentrySTDinterwordspacing{\spaceskip=0pt\relax}
\providecommand\BIBentryALTinterwordstretchfactor{4}
\providecommand\BIBentryALTinterwordspacing{\spaceskip=\fontdimen2\font plus
\BIBentryALTinterwordstretchfactor\fontdimen3\font minus \fontdimen4\font\relax}
\providecommand\BIBforeignlanguage[2]{{%
\expandafter\ifx\csname l@#1\endcsname\relax
\typeout{** WARNING: IEEEtran.bst: No hyphenation pattern has been}%
\typeout{** loaded for the language `#1'. Using the pattern for}%
\typeout{** the default language instead.}%
\else
\language=\csname l@#1\endcsname
\fi
#2}}

\bibitem{Hilmarsen2025}
H.~Hilmarsen, \emph{et~al.}, ``{Visual Lidar Simple Online and Real-Time Tracking (VLSORT)},'' in \emph{ICRA, to be published}, 2025.

\bibitem{Wilthil2017}
E.~F. Wilthil, A.~L. Flåten, and E.~F. Brekke, ``\BIBforeignlanguage{en}{A {Target} {Tracking} {System} for {ASV} {Collision} {Avoidance} {Based} on the {PDAF}},'' in \emph{\BIBforeignlanguage{en}{Sensing and Control for Autonomous Vehicles}}, ser. Lecture {Notes} in {Control} and {Information} {Sciences}, T.~I. Fossen, K.~Y. Pettersen, and H.~Nijmeijer, Eds.\hskip 1em plus 0.5em minus 0.4em\relax Cham: Springer International Publishing, 2017, pp. 269--288.

\bibitem{Wilthil2018Track}
E.~F. Wilthil, E.~Brekke, and O.~B. Asplin, ``Track initiation for maritime radar tracking with and without prior information,'' in \emph{{FUSION}}, July 2018, pp. 1--8.

\bibitem{Helgesen2022}
{\O}.~K. Helgesen, \emph{et~al.}, ``\BIBforeignlanguage{en}{Heterogeneous multi-sensor tracking for an autonomous surface vehicle in a littoral environment},'' \emph{\BIBforeignlanguage{en}{Ocean Engineering}}, Vol. 252, p. 111168, May 2022.

\bibitem{Chen2022a}
X.~Chen, \emph{et~al.}, ``Automatic {Labeling} to {Generate} {Training} {Data} for {Online} {LiDAR}-{Based} {Moving} {Object} {Segmentation},'' \emph{IEEE Robotics and Automation Letters}, Vol.~7, No.~3, pp. 6107--6114, July 2022.

\bibitem{Lim2021}
H.~Lim, S.~Hwang, and H.~Myung, ``{ERASOR}: {Egocentric} {Ratio} of {Pseudo} {Occupancy}-{Based} {Dynamic} {Object} {Removal} for {Static} {3D} {Point} {Cloud} {Map} {Building},'' \emph{IEEE Robotics and Automation Letters}, Vol.~6, No.~2, pp. 2272--2279, Apr. 2021.

\bibitem{Kim2020}
G.~Kim and A.~Kim, ``Remove, then {Revert}: {Static} {Point} cloud {Map} {Construction} using {Multiresolution} {Range} {Images},'' in \emph{{IROS}}, Oct. 2020, pp. 10\,758--10\,765.

\bibitem{Arora2021}
M.~Arora, \emph{et~al.}, ``Mapping the {Static} {Parts} of {Dynamic} {Scenes} from {3D} {LiDAR} {Point} {Clouds} {Exploiting} {Ground} {Segmentation},'' in \emph{{ECMR}}, Aug. 2021, pp. 1--6.

\bibitem{Pfreundschuh2021}
P.~Pfreundschuh, \emph{et~al.}, ``Dynamic {Object} {Aware} {LiDAR} {SLAM} based on {Automatic} {Generation} of {Training} {Data},'' in \emph{{ICRA}}, May 2021, pp. 11\,641--11\,647.

\bibitem{Jinno2019}
I.~Jinno, Y.~Sasaki, and H.~Mizoguchi, ``{3D} {Map} {Update} in {Human} {Environment} {Using} {Change} {Detection} from {LIDAR} {Equipped} {Mobile} {Robot},'' in \emph{{SII}}, Jan. 2019, pp. 330--335.

\bibitem{Ding2018}
X.~Ding, \emph{et~al.}, ``Multi-{Session} {Map} {Construction} in {Outdoor} {Dynamic} {Environment},'' in \emph{{RCAR}}, Aug. 2018, pp. 384--389.

\bibitem{Pomerleau2014}
F.~Pomerleau, \emph{et~al.}, ``Long-term {3D} map maintenance in dynamic environments,'' in \emph{{ICRA}}, May 2014, pp. 3712--3719.

\bibitem{Mersch2023}
B.~Mersch, \emph{et~al.}, ``Building {Volumetric} {Beliefs} for {Dynamic} {Environments} {Exploiting} {Map}-{Based} {Moving} {Object} {Segmentation},'' \emph{IEEE Robotics and Automation Letters}, Vol.~8, No.~8, pp. 5180--5187, Aug. 2023.

\bibitem{Hosseinyalamdary2015}
S.~Hosseinyalamdary, Y.~Balazadegan, and C.~Toth, ``\BIBforeignlanguage{en}{Tracking {3D} {Moving} {Objects} {Based} on {GPS}/{IMU} {Navigation} {Solution}, {Laser} {Scanner} {Point} {Cloud} and {GIS} {Data}},'' \emph{\BIBforeignlanguage{en}{ISPRS International Journal of Geo-Information}}, Vol.~4, No.~3, pp. 1301--1316, Sept. 2015.

\bibitem{Wang2003}
C.-C. Wang, C.~Thorpe, and S.~Thrun, ``Online simultaneous localization and mapping with detection and tracking of moving objects: theory and results from a ground vehicle in crowded urban areas,'' in \emph{{ICRA}}, Vol.~1, Sept. 2003, pp. 842--849 vol.1.

\bibitem{Miyasaka2009}
T.~Miyasaka, Y.~Ohama, and Y.~Ninomiya, ``Ego-motion estimation and moving object tracking using multi-layer {LIDAR},'' in \emph{{IV}}, June 2009, pp. 151--156.

\bibitem{Thompson2019}
D.~Thompson, E.~Coyle, and J.~Brown, ``Efficient {LiDAR}-{Based} {Object} {Segmentation} and {Mapping} for {Maritime} {Environments},'' \emph{IEEE Journal of Oceanic Engineering}, Vol.~44, No.~2, pp. 352--362, Apr. 2019.

\bibitem{Xie2023}
X.~Xie, H.~Wei, and Y.~Yang, ``\BIBforeignlanguage{en}{Real-{Time} {LiDAR} {Point}-{Cloud} {Moving} {Object} {Segmentation} for {Autonomous} {Driving}},'' \emph{\BIBforeignlanguage{en}{Sensors}}, Vol.~23, No.~1, p. 547, Jan. 2023.

\bibitem{Chen2021a}
X.~Chen, \emph{et~al.}, ``Moving {Object} {Segmentation} in {3D} {LiDAR} {Data}: {A} {Learning}-{Based} {Approach} {Exploiting} {Sequential} {Data},'' \emph{IEEE Robotics and Automation Letters}, Vol.~6, No.~4, pp. 6529--6536, Oct. 2021.

\bibitem{Postica2016}
G.~Postica, A.~Romanoni, and M.~Matteucci, ``Robust moving objects detection in lidar data exploiting visual cues,'' in \emph{{IROS}}, Oct. 2016, pp. 1093--1098.

\bibitem{Yan2014}
J.~Yan, \emph{et~al.}, ``Automatic {Extraction} of {Moving} {Objects} from {Image} and {LIDAR} {Sequences},'' in \emph{{3DV}}, Vol.~1, Dec. 2014, pp. 673--680.

\bibitem{Sun2024}
S.~Sun, W.~Guan, and Y.~Wang, ``Occupancy grid based environment sensing for {MASS} in complex waters,'' \emph{Ocean Engineering}, Vol. 303, p. 117848, July 2024.

\bibitem{Dewan2016a}
A.~Dewan, \emph{et~al.}, ``Motion-based detection and tracking in {3D} {LiDAR} scans,'' in \emph{{ICRA}}, May 2016, pp. 4508--4513.

\bibitem{Dewan2016}
------, ``Rigid scene flow for {3D} {LiDAR} scans,'' in \emph{{IROS}}, Oct. 2016, pp. 1765--1770.

\bibitem{Lenz2011}
P.~Lenz, \emph{et~al.}, ``Sparse scene flow segmentation for moving object detection in urban environments,'' in \emph{{IV}}, June 2011, pp. 926--932.

\bibitem{Lee2010}
K.~W. Lee, \emph{et~al.}, ``Tracking random finite objects using {3D}-{LIDAR} in marine environments,'' in \emph{{SAC}}.\hskip 1em plus 0.5em minus 0.4em\relax New York, NY, USA: Association for Computing Machinery, Mar. 2010, pp. 1282--1287.

\bibitem{Nuss2018}
D.~Nuss, \emph{et~al.}, ``\BIBforeignlanguage{en}{A random finite set approach for dynamic occupancy grid maps with real-time application},'' \emph{\BIBforeignlanguage{en}{The International Journal of Robotics Research}}, Vol.~37, No.~8, pp. 841--866, July 2018.

\bibitem{Negre2014}
A.~Nègre, L.~Rummelhard, and C.~Laugier, ``Hybrid sampling {Bayesian} {Occupancy} {Filter},'' in \emph{{IV}}, June 2014, pp. 1307--1312.

\bibitem{Tanzmeister2014}
G.~Tanzmeister, \emph{et~al.}, ``Grid-based mapping and tracking in dynamic environments using a uniform evidential environment representation,'' in \emph{{ICRA}}, May 2014, pp. 6090--6095.

\bibitem{Schreiber2021}
M.~Schreiber, \emph{et~al.}, ``Dynamic {Occupancy} {Grid} {Mapping} with {Recurrent} {Neural} {Networks},'' in \emph{{ICRA}}, May 2021, pp. 6717--6724.

\bibitem{Vatavu2020}
A.~Vatavu, \emph{et~al.}, ``From {Particles} to {Self}-{Localizing} {Tracklets}: {A} {Multilayer} {Particle} {Filter}-{Based} {Estimation} for {Dynamic} {Grid} {Maps},'' \emph{IEEE Intelligent Transportation Systems Magazine}, Vol.~12, No.~4, pp. 149--168, 2020.

\bibitem{Bibby2010}
C.~Bibby and I.~Reid, ``A hybrid {SLAM} representation for dynamic marine environments,'' in \emph{{ICRA}}, May 2010, pp. 257--264.

\bibitem{Vu2011}
T.-D. Vu, J.~Burlet, and O.~Aycard, ``Grid-based localization and local mapping with moving object detection and tracking,'' \emph{Information Fusion}, Vol.~12, No.~1, pp. 58--69, Jan. 2011.

\bibitem{Gies2018}
F.~Gies, A.~Danzer, and K.~Dietmayer, ``Environment {Perception} {Framework} {Fusing} {Multi}-{Object} {Tracking}, {Dynamic} {Occupancy} {Grid} {Maps} and {Digital} {Maps},'' in \emph{{ITSC}}, Nov. 2018, pp. 3859--3865.

\bibitem{Pieper2024}
F.~Pieper and A.~Hahn, ``A conceptual approach to harbor object detection: The potential of 3d-lidar-based sensor fusion for high precision enc,'' in \emph{IFAC CAMS}, 2024.

\bibitem{Obradovic2024}
J.~Obradović, \emph{et~al.}, ``Analysis of lidar-camera fusion for marine situational awareness with emphasis on cluster selection in camera frustum,'' in \emph{IFAC CAMS}, 2024.

\bibitem{Yao2023}
Z.~Yao, \emph{et~al.}, ``{LiDAR}-based simultaneous multi-object tracking and static mapping in nearshore scenario,'' \emph{Ocean Engineering}, Vol. 272, p. 113939, Mar. 2023.

\bibitem{Lin2022}
J.~Lin, \emph{et~al.}, ``Maritime {Environment} {Perception} {Based} on {Deep} {Learning},'' \emph{IEEE Transactions on Intelligent Transportation Systems}, Vol.~23, No.~9, pp. 15\,487--15\,497, Sept. 2022.

\bibitem{Chi2024}
P.~Chi, \emph{et~al.}, ``\BIBforeignlanguage{en}{Online static point cloud map construction based on {3D} point clouds and {2D} images},'' \emph{\BIBforeignlanguage{en}{The Visual Computer}}, Vol.~40, No.~4, pp. 2889--2904, Apr. 2024.

\bibitem{Wang2023}
C.-Y. Wang, A.~Bochkovskiy, and H.-Y.~M. Liao, ``{YOLOv7}: {Trainable} bag-of-freebies sets new state-of-the-art for real-time object detectors,'' in \emph{CVPR}, June 2023, pp. 7464--7475.

\bibitem{forstner2016photogrammetric}
W.~F{\"o}rstner and B.~P. Wrobel, \emph{Photogrammetric computer vision}.\hskip 1em plus 0.5em minus 0.4em\relax Springer, 2016.

\bibitem{ester1996density}
M.~Ester, \emph{et~al.}, ``A density-based algorithm for discovering clusters in large spatial databases with noise,'' in \emph{KDD}, Vol.~96, 1996, pp. 226--231.

\bibitem{Brekke2021}
E.~F. Brekke, A.~G. Hem, and L.-C.~N. Tokle, ``Multitarget {Tracking} {With} {Multiple} {Models} and {Visibility}: {Derivation} and {Verification} on {Maritime} {Radar} {Data},'' \emph{IEEE Journal of Oceanic Engineering}, Vol.~46, No.~4, pp. 1272--1287, Oct. 2021.

\bibitem{Jocher_YOLO_by_Ultralytics_2023}
G.~Jocher, A.~Chaurasia, and J.~Qiu, ``{YOLO by Ultralytics},'' Jan. 2023, YOLOv8 tool GitHub code repository.

\bibitem{Lin2014}
T.-Y. Lin, \emph{et~al.}, ``\BIBforeignlanguage{en}{Microsoft {COCO}: {Common} {Objects} in {Context}},'' in \emph{\BIBforeignlanguage{en}{ECCV}}, D.~Fleet, \emph{et~al.}, Eds.\hskip 1em plus 0.5em minus 0.4em\relax Cham: Springer International Publishing, 2014, pp. 740--755.

\bibitem{Sun2023}
Z.~Sun, \emph{et~al.}, ``Marine ship instance segmentation by deep neural networks using a global and local attention ({GALA}) mechanism,'' \emph{PLOS ONE}, Vol.~18, No.~2, Feb. 2023.

\bibitem{Iancu2021}
B.~Iancu, \emph{et~al.}, ``\BIBforeignlanguage{en}{{ABOships}—{An} {Inshore} and {Offshore} {Maritime} {Vessel} {Detection} {Dataset} with {Precise} {Annotations}},'' \emph{\BIBforeignlanguage{en}{Remote Sensing}}, Vol.~13, No.~5, p. 988, Jan. 2021.

\bibitem{OpenStreetMap}
{OSM contributors}, ``{Map polygons retrieved using OSMnx},'' \url{ https://www.openstreetmap.org }, 2017.

\bibitem{boeing2024modeling}
G.~Boeing, ``Modeling and analyzing urban networks and amenities with osmnx,'' 2024.

\end{thebibliography}

\end{document}